\documentclass[letterpaper, 10 pt, conference]{ieeeconf}  

\IEEEoverridecommandlockouts                              

\usepackage{diagbox}
\usepackage{slashbox}
\usepackage[pdftex]{graphicx} 
\usepackage{grffile}

\newcommand{\mypar}[1]{\vspace{0.1cm}\noindent\textbf{#1}.}
\usepackage{colortbl}
\usepackage[usenames,dvipsnames]{xcolor}
\usepackage{booktabs}
\usepackage{multirow}
\usepackage{float}
\usepackage[hypcap=false,font=footnotesize]{caption}
\usepackage{graphicx}
\usepackage[section]{placeins}
\usepackage{etoolbox,lipsum}
\usepackage{afterpage}
\usepackage{booktabs}
\usepackage{subcaption}
\usepackage{threeparttable}
\usepackage{capt-of,etoolbox}
\usepackage[section]{placeins}
\usepackage{etoolbox,lipsum}
\usepackage{multirow, makecell}
\usepackage{multirow}
\usepackage{gensymb}
\usepackage[usenames,dvipsnames]{xcolor}
\usepackage{soul}
\usepackage{color}
\newcommand{\red}[1]{\textcolor{red}{#1}}
\usepackage{expl3}
\usepackage{svg}
\ExplSyntaxOn
\newcommand\latinabbrev[1]{
	\peek_meaning:NTF . {
		#1\@}%
	{ \peek_catcode:NTF a {
			#1.\@ }%
		{#1.\@}}}
\ExplSyntaxOff
\def\eg{\latinabbrev{e.g}}

\usepackage{txfonts}

\usepackage{tikz}

\newcommand\copyrighttext{%
	\footnotesize Accepted at IROS 2022, \copyright IEEE. Personal use is permitted, but republication/redistribution requires IEEE permission.  Permission from IEEE must be obtained for all other uses, in any current or future media,including reprinting/republishing this material for advertising or promotional purposes, creating new collective works, for resale or redistribution to servers or lists, or reuse of any copyrighted component of this work in other works.}

\newcommand\copyrightnotice{%
	\begin{tikzpicture}[remember picture,overlay]
	\node[anchor=south,yshift=10pt] at (current page.south) {\fbox{\parbox{\dimexpr\textwidth-\fboxsep-\fboxrule\relax}{\copyrighttext}}};
	\end{tikzpicture}%
}

\usepackage{balance}
\usepackage{stfloats}

\definecolor{lightsteelblue}{RGB}{176,196,222}
\definecolor{lightsteelred}{RGB}{230,176,160}
\definecolor{lightsteellila}{RGB}{175,181,224}
\definecolor{lightsteelgreen}{RGB}{182,214,207}
\definecolor{lightsteelyellow}{RGB}{240,240,160}
\definecolor{lightsteelwhite}{RGB}{255,255,255}
\definecolor{lightsteelgray}{RGB}{205,201,201}
\definecolor{lightsteellightgray}{RGB}{210,210,210}

\definecolor{pp_blue}{RGB}{68,114,196}
\definecolor{pp_orange}{RGB}{237,125,49}
\definecolor{pp_lila}{RGB}{112,48,160}
\definecolor{pp_ygreen}{RGB}{112,173,71}

\definecolor{gg_blue}{RGB}{52,138,189}
\definecolor{gg_red}{RGB}{226,74,51}
\definecolor{gg_lila}{RGB}{152,142,213}
\definecolor{gg_green}{RGB}{112,173,71}
\definecolor{under70}{RGB}{255, 255, 255}
\definecolor{overone}{RGB}{192,196,192}
\definecolor{table_standard}{RGB}{230,153,0}
\definecolor{table_uncertainty}{RGB}{112,48,160}

\usepackage{tikz}
\usepackage{pifont} 
\usepackage[cp1251]{inputenc}
\def\endthebibliography{%
	\def\@noitemerr{\@latex@warning{Empty `thebibliography' environment}}%
	\endlist
}

\usepackage{lipsum}

\makeatletter
\let\NAT@parse\undefined
\makeatother

\usepackage{hyperref}
\hypersetup{
    colorlinks=true,
    linkcolor=black,
    filecolor=black,      
    urlcolor=magenta,
    pdftitle={Overleaf Example},
    pdfpagemode=FullScreen,
    }

\newcommand{\deltap}[1]{{\color{OliveGreen}(+#1)}}
\newcommand{\deltam}[1]{{\color{Maroon}(-#1)}}

\newcommand*\OK{\ding{51}}

\title{\LARGE \bf  Multimodal  Generation of Novel Action Appearances for Synthetic-to-Real Recognition of Activities of Daily Living}

\author{Zdravko Marinov$^\star$ \quad \quad \quad David Schneider$^\star$ \quad \quad \quad Alina Roitberg$^\star$ \quad \quad \quad Rainer Stiefelhagen
\\ \\Institute for Anthropomatics and Robotics
\\ Karlsruhe Institute of Technology, Germany
\\  {\tt\scriptsize \quad \quad zdravko.marinov@kit.edu \quad \quad david.schneider@kit.edu \quad \quad \quad  \quad alina.roitberg@kit.edu \quad \quad rainer.stiefelhagen@kit.edu}
\thanks{$^\star$ denotes equal contribution}%
}

\begin{document}

\maketitle
\copyrightnotice{}
\thispagestyle{empty}
\pagestyle{empty}


\begin{abstract}
Domain shifts, such as appearance changes, are a key challenge in real-world applications of activity recognition models, which range from assistive robotics and smart homes to driver observation in intelligent vehicles.
For example, while simulations are an excellent way of economical data collection, a \textsc{Synthetic$\rightarrow$ Real} domain shift leads to $>\textbf{ 60}\%$ drop in accuracy when recognizing Activities of Daily Living (ADLs).


We tackle this challenge and introduce an activity domain generation framework which creates novel ADL appearances (\textit{novel} domains) from different existing activity modalities (\textit{source} domains) inferred from video training data. 
Our framework computes human poses, heatmaps of body joints, and optical flow maps and uses them alongside the original RGB videos to learn the essence of source domains in order to generate completely new ADL domains. 
The model is optimized by \emph{maximizing} the distance between the existing source appearances and the generated novel appearances while ensuring that the semantics of an activity is preserved through an additional classification loss.
While source data multimodality is an important concept in this design, our setup does not rely on multi-sensor setups, (i.e., all source modalities are inferred from a single video only.)
The newly created activity domains are then integrated in the training of the ADL classification networks, resulting in models far less susceptible to changes in data distributions.
Extensive experiments on the \textsc{Synthetic$\rightarrow$Real} benchmark Sims4Action demonstrate the potential of the domain generation paradigm for cross-domain ADL recognition, setting new state-of-the-art results.
Our code is publicly available at \url{https://github.com/Zrrr1997/syn2real_DG}.

\end{abstract}


\section{Introduction}
When roboticists apply  visual activity recognition models in practice, they will quickly discover the problem of domain shifts.
In fact, a model is rarely deployed under conditions identical to the ones in the training set, as we face changes in illumination, camera type and -placement~\cite{wulfmeier2017addressing, james2019sim}. 
One domain change vital in robotic ADL assistance is the transition from synthetic to real data, as simulations ease the burden of intrusive data collection and privacy concerns in domestic environments~\cite{li2021igibson, RoitbergSchneider2021Sims4ADL, james2019sim, Hwang2021}.
Especially in the light of the ageing population, domain-invariant recognition of Activities of Daily Living (ADL) is an important element for household robot perception and human-tailored planning~\cite{christoforou2019overview,marco2018computer}.

\begin{figure}
    \centering
    \includegraphics[width=\linewidth]{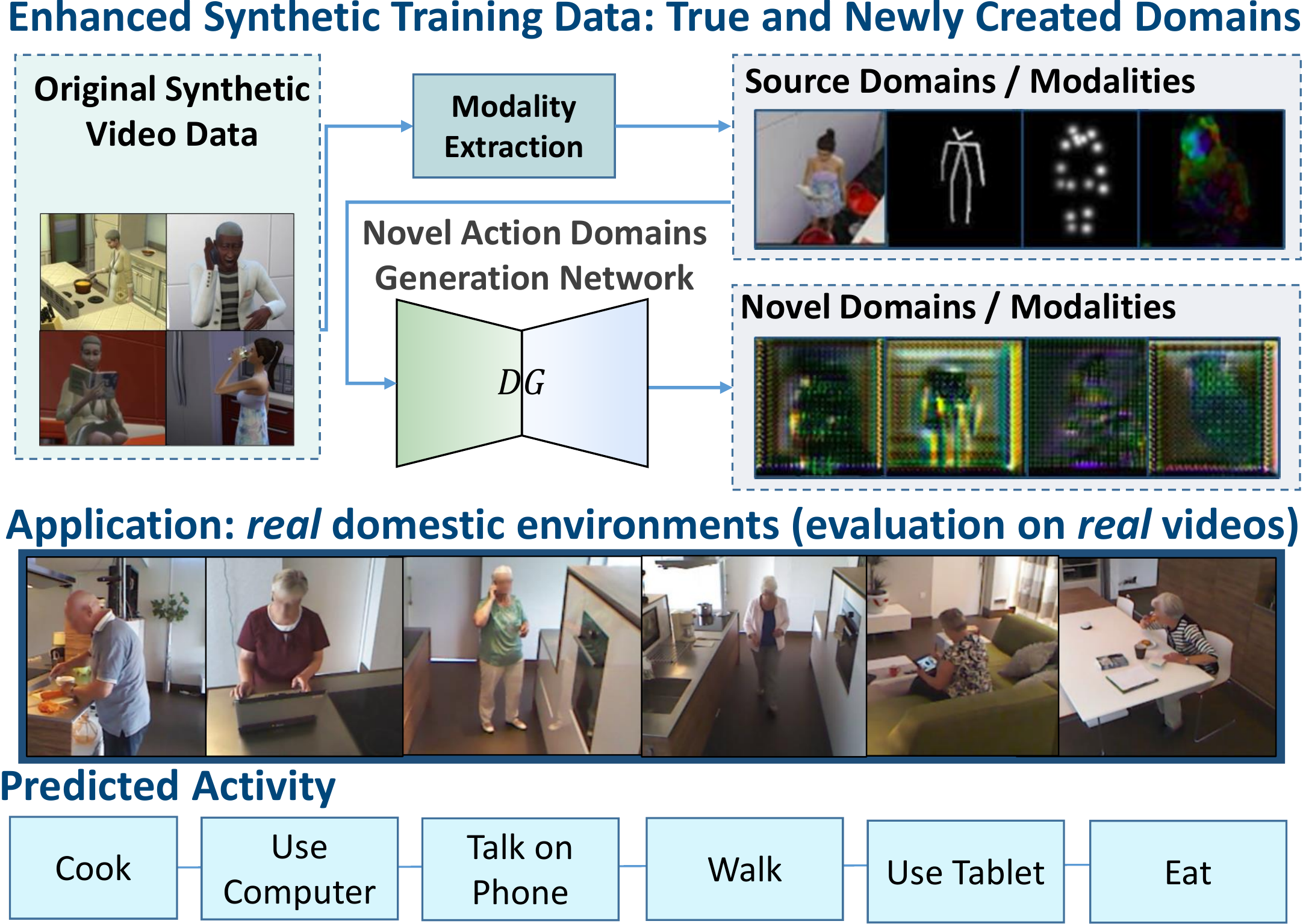}
    \caption{An overview of the proposed  \textsc{Synthetic$\rightarrow$Real} activity recognition pipeline improved through neural network-based generation of novel activity domains. In addition to synthetic RGB videos, the training data is enriched with modalities explicitly extracted from videos (\textit{source} modalities) and action representations from the \textit{novel} modalities generated with our model. Such multimodal diversification of the training samples significantly mitigates adverse effects of the \textsc{Synthetic$\rightarrow$Real} domain shift.}
    \label{fig:intro}
    \vspace{-1.5em}
\end{figure}
\begin{figure*}
    \centering
    \includegraphics[width=0.9\textwidth]{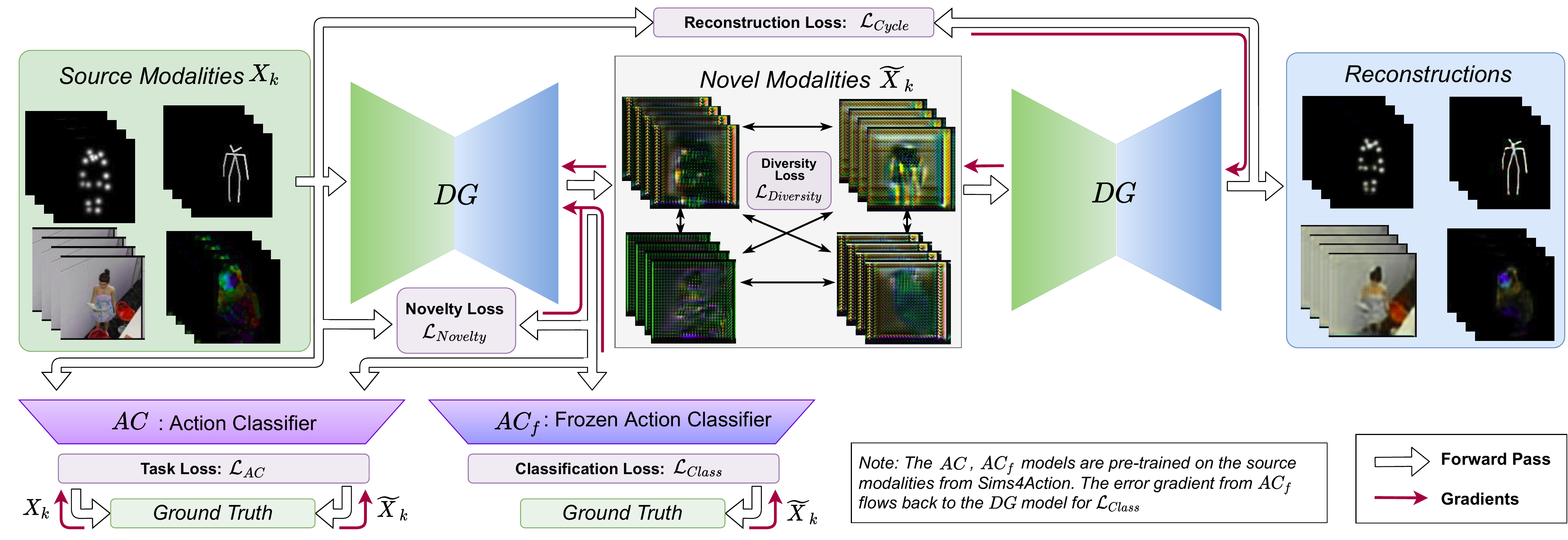}
    \caption{Architecture of our multimodal framework for generation of novel ADL modalities. First, multiple modalities related to the body pose and movement are extracted and used alongside the original RGB videos as our source modalities $X_k$. Activity examples in these source modalities are then fed into the \textit{Domain Generator (DG)} as single images in order to generate novel modalities. The distribution divergence between each source-novel domain pair is estimated via the \textit{Novelty Loss}, whereas the \textit{Diversity Loss} computes the divergence for each novel-novel domain pair. The source domains and the generated novel domains are both used as inputs to the \textit{Action Classifier}, which makes separate predictions for both. The same generator is used to reconstruct the original input and the \textit{Reconstruction Loss} is computed. The error signals for training our model via backpropagation are depicted as red arrows.}
    \vspace{-0.5cm}
    \label{fig:architecture}
\end{figure*}

Recently, the paradigm of novel domain generation has been proposed as a way of extensive data augmentation for image recognition, leading to more domain-invariant models, e.g., for digit classification~\cite{zhou2020learning}. 
At the same time, activity analysis allows a plethora of true modalities based on human movement.
Apart from raw RGB videos, modalities such as optical flow or body poses can be automatically extracted and used as \textit{source} domains for learning to generate more diverse and activity-specific \textit{novel} domains.
Despite its high potential for mitigating domain shifts, such domain generation has been overlooked in activity analysis and is therefore the main motivation of our work.

We aim to make a step towards ADL recognition less susceptible to changes in data distribution and introduce a generative framework enriching the training data through generation of previously unseen activity domains.
In our approach, multimodal action representations derived, \eg, from body pose, joint heatmaps and optical flow, are used to learn creating novel activity domains by \emph{maximizing} the distance between the existing modalities (\textit{source} domains) and the generated appearances (\textit{novel} domains) while ensuring that the semantics of an activity are preserved through an additional \textit{Classification Loss}.
The newly generated novel modalities are then mixed with the initial source modalities and constitute a diversified version of the training set (see overview in Figure \ref{fig:intro}).
Multimodality is a central concept in our framework and we believe that the novel domain generation paradigm suits activity analysis especially well, as body pose and movement dynamics enable a wider range of existing source modalities, which in return encourages higher diversity of the generated novel domains. Nevertheless, our framework does not rely on multi-sensor setups since all source modalities are inferred from RGB videos only.

The problem of domain shift on the \textsc{Synthetic$\rightarrow$Real} benchmark is presented in Table \ref{tab:sota_dg}. The current state-of-the-art model is trained on the synthetic Sims4Action~\cite{RoitbergSchneider2021Sims4ADL} dataset and experiences a large domain gap of $\approx 57\%$ when evaluated on real data. We conduct extensive experiments on the \textsc{Synthetic$\rightarrow$Real} ADL recognition benchmark~\cite{RoitbergSchneider2021Sims4ADL} and make two observations. 
Firstly, multimodality itself is highly beneficial for cross-domain ADL recognition and using the body pose related data representations alone leads to a significant performance boost. 
Secondly, our idea of generating novel ADL appearances for training data enhancement consistently improves recognition results.
Our framework outperforms state-of-the-art results, demonstrating the potential of multimodal domain generation for human activity analysis without using any pre-training on real data.
\begin{table}[!h]
    \centering
    \scalebox{0.85}{
    \begin{tabular}{cl|cc|c}
    \toprule
        \multicolumn{2}{c}{Training: Sims4Action~\cite{RoitbergSchneider2021Sims4ADL}} & \multicolumn{2}{c}{Testing: Balanced Accuracy [\%]} \\ \hline
        Model & Pre-training & \textsc{Synthetic} & \textsc{Real} & Domain Gap \\ \hline
        \multirow{2}{*}{S3D~\cite{xie2018rethinking}}  & Kinetics-400~\cite{kay2017kinetics} &  \ \textbf{84.61}$\red{^\dagger}$ & 23.23 & 61.38 \\ 
        & None & 56.52 & 12.40 & 44.12\\ 
        \multirow{2}{*}{I3D~\cite{carreira2017quo}} & Kinetics-400~\cite{kay2017kinetics} & 81.12 & \  \textbf{23.25}$\red{^*}$ & 57.87 \\ 
        & None & 66.91 & 10.91 & 56.00\\ \hline
        \bottomrule
    \end{tabular}}
    \caption{Current state-of-the-art results for the \textsc{Synthetic $\rightarrow$ Real}$\red{^*}$ and \textsc{Synthetic $\rightarrow$ Synthetic}$\red{^\dagger}$ benchmarks from \textit{Let's Play for Action} \cite{RoitbergSchneider2021Sims4ADL}. All four models are trained on Sims4Action. The \textsc{Synthetic} and \textsc{Real} test sets are Sims4Action~\cite{RoitbergSchneider2021Sims4ADL} and Toyota Smarthome~\cite{das2019toyota} respectively.}
    \label{tab:sota_dg}
    \vspace{-1.5em}
\end{table}

\noindent\textbf{Note on our Terminology.} Similarly to~\cite{zhou2021domain,wang2022generalizing}, we define a \textit{domain} as a joint distribution $P_{XY}$ over a feature space $\mathcal{X}$ and a label space $\mathcal{Y}$. In our work, we extract multiple \textit{modalities} from synthetic data and learn to generate novel \textit{modalities}. Since each modality occupies a distinct feature subspace $\mathcal{X^{'}}$  and exhibits a unique appearance, we use the terms \textbf{modality} and \textbf{domain} interchangeably.

\section{Related Work}


\subsection{Recognizing Activities of Daily Living}
To effectively interact with people, robots need to accurately perceive the current state of the human. 
Despite the impressive progress in general activity classification~\cite{shahroudy2016ntu, kay2017kinetics, smaira2020short,Wang2019, xie2018rethinking, Tran2018,liu2016spatio, hara2017learning} and a variety of frameworks introduced specifically for robotics applications~\cite{christoforou2019overview, jang2020etriactivity3d,roitberg2015multimodal,dreher2019learning, mofijul2020hamlet, roitberg2014human}, this task remains very challenging in robotics, as agents often operate in a dynamic world where changes in concept-of-interest and data appearances may occur at any time~\cite{sunderhauf2018limits}.
In assistive robotics, recognizing Activities of Daily Living (ADL) is especially interesting and is often addressed by collecting and labelling new datasets tailored for the ADLs- and environments-of-interest~\cite{sigurdsson2016hollywood, damen2018scaling, das2019toyota, jang2020etriactivity3d}.
Creating such datasets which intend to realistically reflect real-world households 
requires larger efforts for sensory setups and data curation which results in  datasets being smaller in comparison to general action classification benchmarks often created from web data \cite{kay2017kinetics,caba2015activitynet}. 
Methodologically, ADL recognition research is strongly influenced by architectures introduced in general video classification, with 3D Convolutional Neural Networks (CNNs)~\cite{carreira2017quo, xie2018rethinking,tran2015learning} being common backbone architectures~\cite{das2019toyota,dai2020toyota}, but also more specialized approaches often derived from the body pose, have been introduced~\cite{das2020vpn, das2021vpn, mofijul2020hamlet}.
At the same time, recent research has raised alarming evidence, that deep learning-based ADL recognition approaches are very sensitive to changes in data distribution~\cite{RoitbergSchneider2021Sims4ADL}.
Mitigating this effect by exploring the domain generation paradigm~\cite{zhou2020learning} in the field of ADL recognition for the first time is the main contribution of our work.

\subsection{Synthetic Human Actions}
Given the difficulty of collecting labeled datasets, learning from simulated data has been researched in many different fields of computer vision and is also emerging in video-based learning tasks, for example for pose recognition~\cite{shotton2011real, varol2017learning, liu2019learning} and more recently in the domain of human activity recognition \cite{de2017procedural, puig2018virtualhome, RoitbergSchneider2021Sims4ADL, ludl2020enhancing, Hwang2021}. The latter works focus on either augmenting existing training data by mixing it with generated data \cite{de2017procedural, zhang2019rsa, Hwang2021}, learning action categories on synthetic data only \cite{RoitbergSchneider2021Sims4ADL} or on learning compositions of actions within virtual domains \cite{puig2018virtualhome}. \cite{varol2021synthetic} make use of a hybrid approach and combine real videos with rendered synthetic humans shown from different viewpoints.
While synthetic examples are an excellent alternative to intrusive and time-consuming ADL dataset creation, the transition from simulations to real data at test-time comes with a remarkable performance drop~\cite{RoitbergSchneider2021Sims4ADL} (see Table \ref{tab:sota_dg}). In this work, we focus on the \textsc{Synthetic$\rightarrow$Real} transition in ADL recognition. We introduce a multimodal framework which leads to more domain-invariant recognition models by learning to generate new appearance versions of the synthetic training samples and by using them to diversify the training dataset.

\subsection{Domain Generalization and Adaptation}
Unsupervised domain adaptation methods learn a task on a source domain and try to solve this task on a target domain by learning a mapping given unlabelled data from the target domain, a task which has seen significant development in recent years in the field of video-based learning~\cite{choi2020shuffle, chen2019temporal, chen2020action, busto2018open, jamal2018deep, choi2020unsupervised, pan2020adversarial, reiss2020deep}.
In contrast, domain generalization describes the ability to maintain performance on a target domain despite not having access to any training data from this domain. 
Recently, Zhou et al.~\cite{zhou2020learning} proposed the domain generation paradigm, which is fundamentally different from previous work, as it learns to map source data to \textit{unseen}, \textit{newly generated} domains. 
Our work extends the image-based technique of \cite{zhou2020learning}, for the first time exploring it in the scope of ADL recognition and video recognition in general, which opens many additional possibilities of ADL-related source modalities, such as body poses or optical flow. 


%

\section{Multimodal Generation of  Activity Domains}
\label{sec:approach}
\begin{figure}
    \centering
    \includegraphics[width=0.4\textwidth]{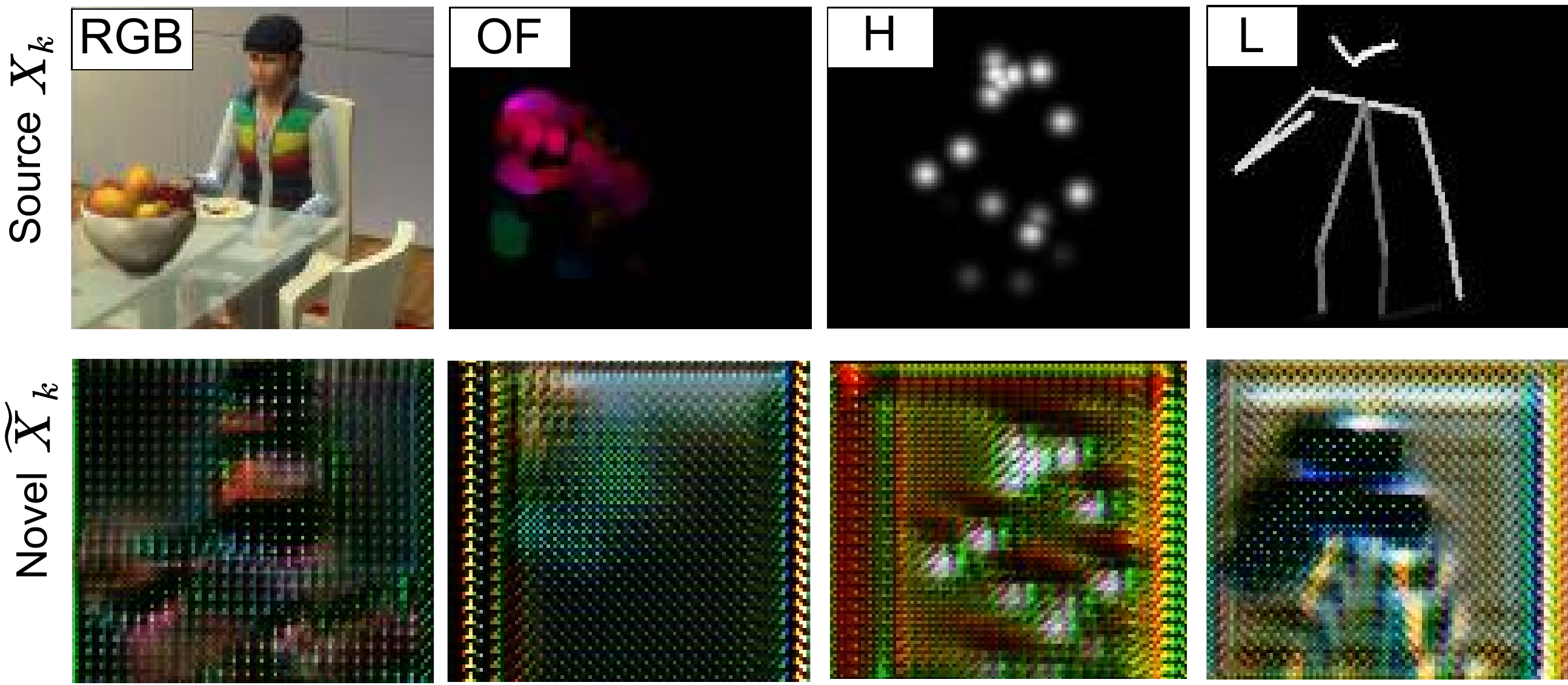}
    \caption{Examples of the \textit{source} $X_k$ (top row) and \textit{novel} modalities $\widetilde{X}_k$ (bottom row). H: Heatmaps, L: Limbs, OF: Optical Flow. We use all 8 modalities $X_k \cup \widetilde{X}_k$ for training and evaluate our models on \textsc{Real} data.}
    \vspace{-0.5cm}
    \label{fig:modality_examples}
\end{figure}


In this section, we introduce a multimodal framework for better domain generalization in activity recognition, aiming to lighten the impact of appearance changes when moving from synthetic ADL training data to real-world robotic applications.
We follow the  domain generation paradigm~\cite{zhou2020learning} recently proposed for image recognition, extending it to the scope of human activity analysis.  
Conceptually, our framework produces \textit{novel} modalities $\widetilde{X}_k$ by learning to transform given \textit{source} modalities $X_k$ and comprises of four main building blocks: 1) A modality extraction module used to compute multiple source modalities from the original RGB video 2) a pre-trained domain classifier $DC$, 3) an action classifier $AC$, 4) and a domain generator $DG$. As we are using each extracted modality as a source \textit{domain} we refer to the source and novel domains as source and novel \textit{modalities}.

The task we address is  \textsc{Synthetic$\rightarrow$Real} domain generalization.
We train an action classifier $AC$ on samples from \textsc{Synthetic} domains $x_s \in X_k \cup \widetilde{X}_k$ with action labels $y \in Y$.
In domain generalization, training and test data originate from distinct probability distributions, in our case $x_s\sim p_{synthetic}$ and $x_r\sim p_{real}$, and test samples $x_r$ neither have labels, nor are seen during training. Our goal is to classify each instance $x_r$ in the \textsc{Real} target test domain $X_r$, which has a shared action label set $Y$ with the training set. For this, we utilize the synthetic Sims4Action~\cite{RoitbergSchneider2021Sims4ADL} dataset for training and the real Toyota  Smarthome~\cite{das2019toyota} and ETRI-Activity3D-LivingLab (ETRI)~\cite{jang2020etriactivity3d} as two separate test sets.
\subsection{Extracting Source Modalities}

The nature of activity recognition and video data in general allows us to leverage a wide range of modalities, such as body pose and movement dynamics, which would not be applicable in conventional image classification. 
We utilize four source modalities $X_k, k\in\{0, 1, 2, 3\}$ which are extracted directly from the training data (\textit{i.e.}, RGB videos). 
The source modalities consist of 1) heatmaps of the body joint locations, 2) limbs connecting the joints as lines, 3) dense optical flow extracted between each two frames, and 4) raw RGB images (see top row of Figure \ref{fig:modality_examples}). 
The heatmaps and limbs are extracted using the AlphaPose \cite{fang2017rmpe, li2018crowdpose, xiu2018poseflow} pose detector, which infers $17$ joint locations.
The heatmaps modality $h(x,y)$ at pixel $(x,y)$ is obtained by applying 2D gaussian maps, centered at each joint location $(x_i, y_i)$ and weighted by its confidence $c_i$ as seen in Equation \ref{eq:gaussian_heatmap}.
\begin{equation}
    h(x,y)=exp\bigg(\frac{-((x - x_i)^2 + (y - y_i)^2)}{2\sigma^2}\bigg) \cdot c_i
    \label{eq:gaussian_heatmap}
\end{equation}
The limbs domain is composed by connecting the joints with white lines and weighting each line by the smaller confidence of its endpoints. The optical flow is estimated using the Gunner-Farneback method \cite{farneback2003two}. 
We refer to these modalities as the four \textit{source} modalities $X_k$ (see top row in Figure \ref{fig:modality_examples}).


\begin{figure}[b]
    \centering
    \includegraphics[width=0.4\textwidth]{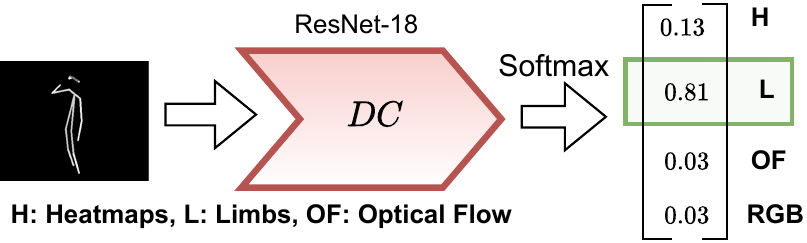}
    \caption{Domain Classifier (DC): Overview of inference and training. The DC model is pre-trained end-to-end on Sims4Action.}
    \label{fig:DC_overview}
    \vspace{-0.8em}
\end{figure}
\subsection{Domain Classifier} 
%

\begin{figure}[t]
    \centering
    \includegraphics[width=0.45\textwidth]{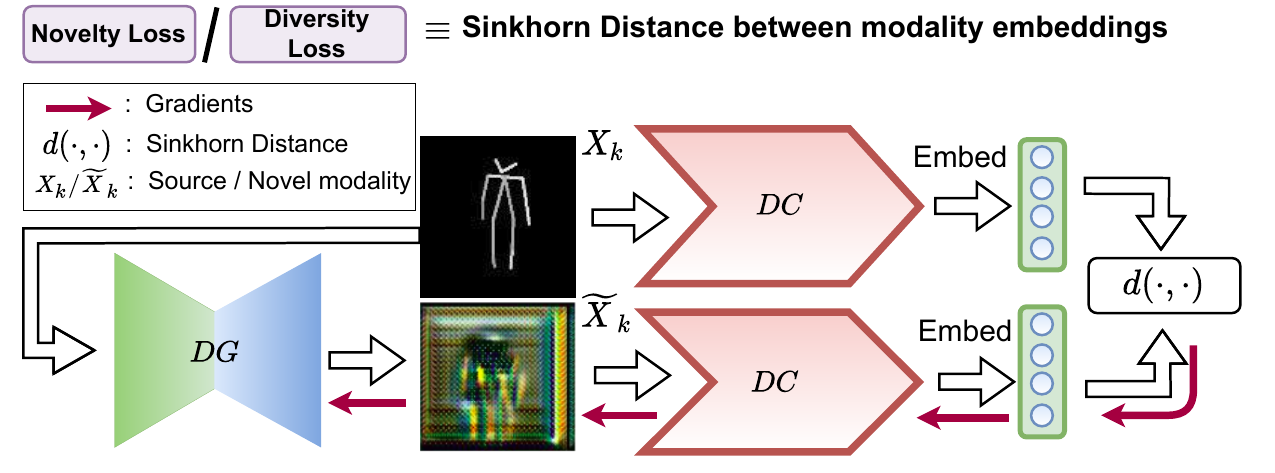}
    \caption{Domain Classifier: Computation of the \textit{Novelty Loss} for all source-novel modality pairs $(X_k, \widetilde{X}_k)$. Note that the \textit{Diversity Loss} is computed the same way with all novel-novel pairs $(\widetilde{X}_k, \widetilde{X}_l)_{k \neq l}$.}
    \label{fig:DC_embed}
    \vspace{-1.5em}
\end{figure}

%
The domain classifier $DC$ is trained on the synthetic Sims4Action~\cite{RoitbergSchneider2021Sims4ADL} to classify the source modalities $X_k$ with labels $k\in\{0, 1, 2, 3\}$ as shown in Figure \ref{fig:DC_overview}.
Afterwards, its weights are frozen and it is utilized for training the domain generator $DG$. The frozen $DC$ is used to obtain embeddings from the source $X_k$ and novel modalities $\widetilde{X}_k$ as seen in Figure \ref{fig:DC_embed}. The Sinkhorn distance~\cite{cuturi2013sinkhorn} between the embeddings is utilized as a distribution divergence metric and it is used to compute the \textit{Novelty Loss} for each source-novel modality pair $(X_k, \widetilde{X}_k)$ and the \textit{Diversity Loss} for all novel-novel pairs $(\widetilde{X}_k, \widetilde{X}_l)_{k \neq l}$ (see Equations \ref{eq:novelty_loss}, \ref{eq:diversity_loss}). The error gradient is propagated back to the domain generator $DG$ and conditions it to produce novel modalities, which are both diverse and different from the source modalities. 
\subsection{Domain Generator}
\label{sec:DG} The goal of the domain generator $DG$ is to extend and diversify the synthetic training data from Sims4Action~\cite{RoitbergSchneider2021Sims4ADL}. The $DG$ model is trained on the four source modalities $X_k$ to generate four novel modalities $DG(X_k) = \widetilde{X}_k$, which should be as diverse as possible, while remaining semantically and structurally consistent. The diversity of the new modalities is enforced by the \textit{Novelty} and \textit{Diversity} loss terms $\mathcal{L}_{Novelty}$ and $\mathcal{L}_{Diversity}$. The \textit{Novelty Loss} maximizes the distribution divergence between the source and novel modalities, while the \textit{Diversity Loss} maximizes the distribution divergence between each pair of generated novel modalities. The divergence measure we use is the Sinkhorn distance~\cite{cuturi2013sinkhorn} $d(\cdot, \cdot)$ between the embeddings obtained by the domain classifier and is computed as illustrated in Figure \ref{fig:DC_embed}. To ensure that the new modalities are dissimilar to the source modalities and are also dissimilar to each other these two loss terms are maximized w.r.t. $DG$ as shown in Equations \ref{eq:novelty_loss} and \ref{eq:diversity_loss}.
\begin{equation}
  \mathcal{L}_{Novelty}= \max_{DG} \sum_{k=0}^3 d(\widetilde{X}_k, X_k) 
    \label{eq:novelty_loss}
\end{equation}
\begin{equation}
    \mathcal{L}_{Diversity}= \max_{DG} \sum_{k=0}^3 \sum_{l=0}^3 d(\widetilde{X}_k, \widetilde{X}_l)
    \label{eq:diversity_loss}
\end{equation}
where $k, l\in\{0,1,2,3\}$, $k\neq l$, and $\widetilde{X}_k=DG(X_k)$.

Furthermore, the domain generator is conditioned to preserve the semantic and structural consistency of the actions in the novel modalities by minimizing the action \textit{Classification Loss} $\mathcal{L}_{Class}$ and the \textit{Reconstruction Loss} $\mathcal{L}_{Cycle}$.
\begin{equation}
          \mathcal{L}_{Class}=\min_{DG} \sum_{k=0}^3 \mathcal{L}_{CE}(AC_f(\widetilde{X}_k), Y_k)   \label{eq:class_loss}
\end{equation}
\begin{equation}
    \mathcal{L}_{Cycle} = \min_{DG}\sum_{k=0}^3||DG(DG(X_k)), X_k||_1 \label{eq:cycle_loss}
\end{equation}
where $Y_k$ is the ground-truth action label, $\mathcal{L}_{CE}$ is the cross-entropy loss, and $AC_f$ is the frozen action classifier pre-trained on Sims4Action's~\cite{RoitbergSchneider2021Sims4ADL} source modalities $X_k$ (explained in Section \ref{subsec:training}). Note that we train on mini-batches. For this reason, $X_k, \widetilde{X}_k, Y_k$ refer to individual batches of size $B$ in all equations, rather than the whole dataset.

\subsection{Action Classifier} \label{subsec:ac}
The action classifier $AC$ is firstly pre-trained only on the source modalities $X_k$ and then trained further on both $X_k$ and $\widetilde{X}_k$ to assign the correct activity label. This encourages the model to learn representations in both the source and novel modalities and results in a larger and a more heterogeneous training set. The novel modalities increase the diversity of the training samples as their distribution differences are maximized by $\mathcal{L}_{Novelty}$ and $\mathcal{L}_{Diversity}$, which leads to a more versatile training dataset.
The generation of $\widetilde{X}_k$ can hence be viewed as a method for data augmentation. 

%
%
\begin{figure}[]
    \centering
    \includegraphics[width=0.4\textwidth]{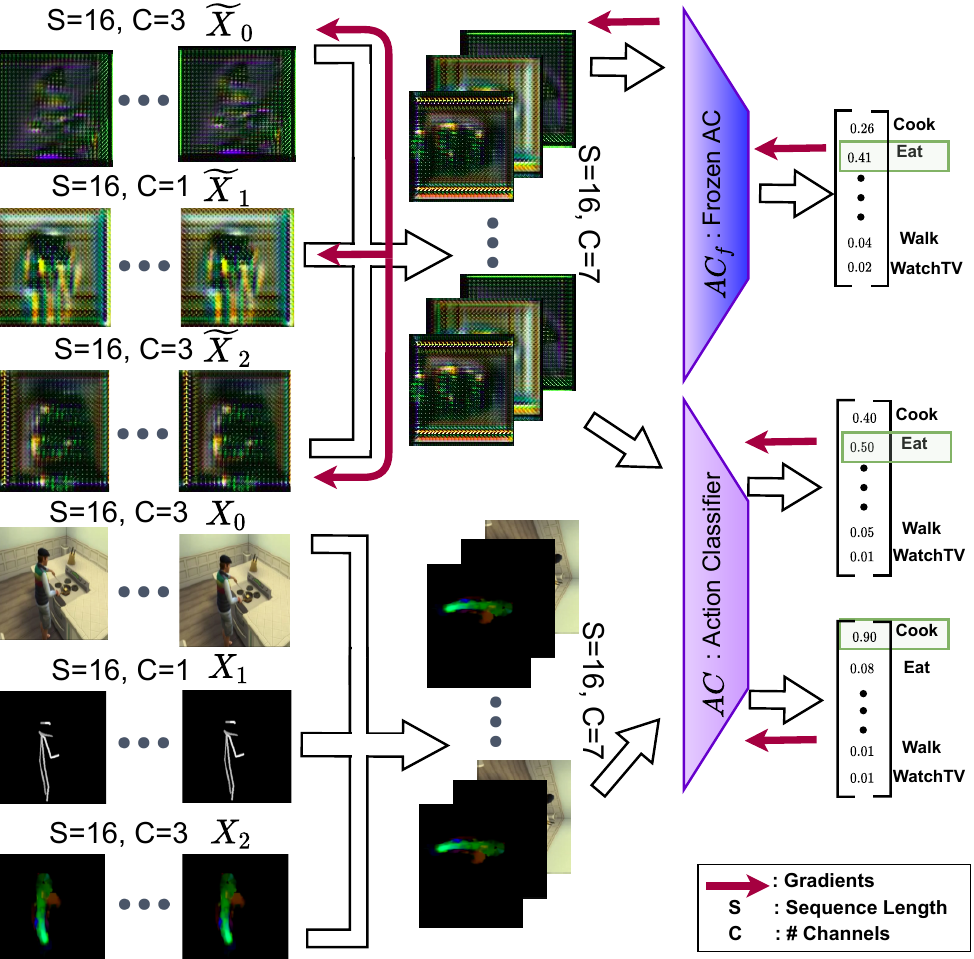}
    \caption{Example of  $AC$ training with three modalities $(k=2)$. The input sequences are concatenated along the channel dimension $C$. The action classifier $AC$ is trained on both the source $X_k$ and novel $\widetilde{X}_k$ modalities for action recognition and makes separate predictions for both. Note that the input to the $AC$ model is sequence-based ($S=16$) as opposed to the image-based $DC$ and $DG$ ($S=1$). Its frozen copy $AC_f$ is pre-trained on the source modalities from Sims4Action and is only used for the computation of $\mathcal{L}_{Class}$ in Equation \ref{eq:class_loss}. The error gradient from $AC_f$ flows back to the novel modalities as its prediction is used to train $DG$ with $\mathcal{L}_{Class}$ (Eq. \ref{eq:class_loss}).}
    \label{fig:AC_overview}
    \vspace{-1em}
\end{figure}
\subsection{Model Architectures}
\label{subsec:arch_training}

As our action classifier $AC$, we utilize a 3D-CNN model to exploit the temporal features in the videos, which are important for action recognition \cite{Hara_2018_CVPR, tran2015learning, varol2017long}. 3D-CNN models such as the Inflated 3D-ConvNet (I3D) \cite{carreira2017quo} have shown remarkable performance in recognition for ADL \cite{das2019toyota, carreira2017quo}. Our action classification 3D-CNN model is implemented with the Separable 3D architecture (S3D) \cite{xie2018rethinking}. The S3D model leverages separable convolutions and replaces most of the 3D convolutions of I3D with cheaper 2D convolutions to reduce the complexity while boosting the performance of I3D. For the domain classifier $DC$ we employ ResNet18 \cite{he2016deep} to learn to classify all source modalities (see Figure \ref{fig:DC_overview}). Lastly, the domain generator $DG$, which also operates on single images, is modelled after Zhou et al. \cite{zhou2020learning} and consists of two down-sampling conv-layers, two residual blocks \cite{he2016deep} with instance normalization \cite{ulyanov2016instance} and two transposed conv-layers to up-sample back to the input's size. 

\subsection{Training Procedure} \label{subsec:training}

All of our models are trained \textit{only} on Sims4Action~\cite{RoitbergSchneider2021Sims4ADL}. The $AC$ model is evaluated in the end on the real Toyota Smarthome~\cite{das2019toyota} and ETRI~\cite{jang2020etriactivity3d} datasets to test its capability for domain generalization. The rest of the models are used only for the computation of loss terms - $AC_f, DC$, or for enhancing $AC$'s training data - $DG$. The domain classifier $DC$ is trained on Sims4Action to distinguish the source modalities $X_k$ as seen in Figure \ref{fig:DC_overview}. Then the $DC$ model is frozen and used solely for the computation of $\mathcal{L}_{Novelty}$ and $\mathcal{L}_{Diversity}$ as in Figure \ref{fig:DC_embed} for the rest of the training. 

Before we jointly train the other two models $AC$ and $DG$, we pre-train an action classifier on all source modalities $X_k$.
We then produce two copies of the pre-trained action classifier - the first copy $AC_f$ is frozen and used only to enforce the semantic consistency of the domain generator in $\mathcal{L}_{Class}$ (see Equation \ref{eq:class_loss}). The second copy $AC$ is further fine-tuned on both the source $X_k$ and novel modalities $\widetilde{X}_k$ with the cross-entropy loss. Only the $AC$ model is evaluated in the end on the real Toyota Smarthome and ETRI datasets. 

\mypar{Loss Computation} In each iteration, a sequence of images is sampled from random chunks of the training videos from all source modalities $X_k$. The modality images from the sequence are concatenated along the channel dimension and are used to train the $AC$ model. Each individual image from the sequence is transformed by $DG$ into the novel modalities $\widetilde{X}_k$, which are reshaped into image sequences (see Figure \ref{fig:AC_overview} top-left). The novel modality sequences are again concatenated along the channels and fed to $AC$ and $AC_f$. The weights of $AC_f$ are not updated, but its error is propagated back to the domain generator to compute $\mathcal{L}_{Class}$ as seen in the red arrows in Figure \ref{fig:AC_overview}. The \textit{Novelty} and \textit{Diversity} loss terms are computed with the
help of the frozen $DC$ model and the Sinkhorn distance~\cite{cuturi2013sinkhorn}
(see Figure \ref{fig:DC_embed}). Finally the \textit{Reconstruction Loss} is
computed by iterating over all source and novel modality
images and applying Equation \ref{eq:cycle_loss}. The final loss function $\mathcal{L}_{DG}$ for the $DG$ model is:
\begin{equation}
    \resizebox{0.44\textwidth}{!}{%
    $\mathcal{L}_{DG}= \displaystyle\min_{DG} \lambda_c\mathcal{L}_{Class} + \lambda_r\mathcal{L}_{Cycle} - \lambda_d(\mathcal{L}_{Novelty} + \mathcal{L}_{Diversity})$
    }
\end{equation}
where $\lambda_c, \lambda_r, \lambda_d$ are balancing parameters for the loss terms.

The $AC$ model is trained on both the source and novel modality sequences. Hence, the \textit{Task Loss} $\mathcal{L}_{AC}$ is:
\begin{equation}
    \resizebox{0.44\textwidth}{!}{%
    $\mathcal{L}_{AC}=\displaystyle\min_{AC} \sum_{k=0}^3 \alpha \mathcal{L}_{CE}(AC(X_k), Y_k) + (1-\alpha) \mathcal{L}_{CE}(AC(\widetilde{X}_k), Y_k)$
    }
\end{equation}
where $\alpha$ balances training on source $X_k$ and novel modalities $\widetilde{X}_k$ and the other terms follow the notation of Equation \ref{eq:class_loss}.

\subsection{Evaluation Procedure} 
Firstly, we evaluate the performance of S3D-based $AC$ models without our domain generalization method. To investigate the effect of multimodality, we train $15$ $AC$ models on all $\sum_{i=1}^4{i \choose 4}=15$ source modality combinations from Sims4Action. We use early fusion via channel concatenation for each of these combinations. Then, we evaluate all $AC$ models on the real Toyota Smarthome~\cite{das2019toyota} and ETRI~\cite{jang2020etriactivity3d} datasets to estimate their domain generalization capabilities.

Afterwards, we apply our domain generalization approach to each of these $AC$ models and create two copies of each - $AC$ and $AC_f$. For each modality combination, we initialize a new domain generator $DG$ and train it alongside its corresponding $AC$ and $AC_f$ models on Sims4Action as described in Section \ref{subsec:training}. A pre-trained domain classifier $DC$ is re-used in all $15$ training sessions. In the end, we evaluate all the $AC$ models on the real Toyota Smarthome~\cite{das2019toyota} and ETRI~\cite{jang2020etriactivity3d} datasets. The motivation for evaluating all modality combinations is to explore which modalities synergize well and to show that the novel domains improve the domain generalization for the vast majority of the 15 modality combinations. 
%
%
%
%
%

\begin{figure}[]
    \centering
    \includegraphics[width=0.90\linewidth]{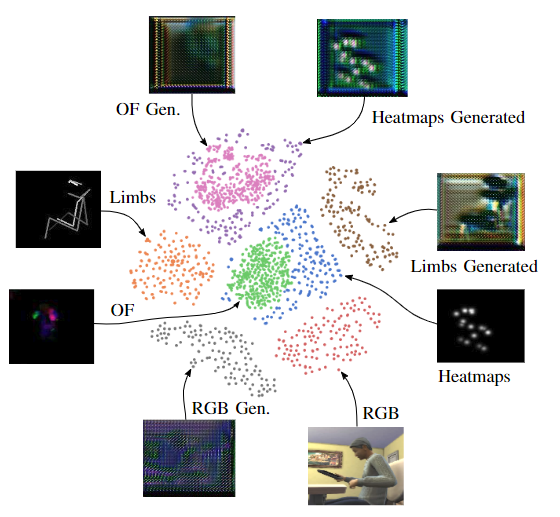}
    
    \caption{Visualizations using t-SNE~\cite{van2008visualizing} embeddings from $DG$ in both, \textit{real and generated} ADL domains. All domains are marked with different colors.}
    \label{fig:tsne}
    \vspace{-1em}
\end{figure}

\subsection{Implementation Details}
We use the same hyperparameters for all of our training sessions and experiments. The action classifiers are pre-trained on Sims4Action's~\cite{RoitbergSchneider2021Sims4ADL} source modalities for 200 epochs. Then the domain generator and action classifier are trained jointly on Sims4Action for 50 epochs as described in Section \ref{subsec:arch_training}. The input frame size is $112 \times 112$ and the sequence length for the S3D action classifier is $S=16$ frames. The input videos are divided into chunks of 90 frames each following the protocol of~\cite{RoitbergSchneider2021Sims4ADL}. During training we randomly sample 16 consecutive frames from the chunks and feed them to the domain generator and action classifier. We utilize the Adam optimizer \cite{kingma2014adam} with $\eta=10^{-4}$, $\beta_1=0.5$, and $\beta_2=0.999$. For regularization, we employ a weight decay of $\lambda=5\cdot10^{-5}$ for all models. We set $\lambda_c=\lambda_d=1$, $\lambda_r=10$, and $\alpha=0.5$ for $\mathcal{L}_{DG}$ and $\mathcal{L}_{AC}$. The embedding size of $DC$ is 512 and we set $\sigma=6$ for Equation \ref{eq:gaussian_heatmap}.
%
%
\section{Experiments}
\label{sec:experiments}
\subsection{Benchmark Details}
We focus on domain generalization between simulated and real data using Sims4Action~\cite{RoitbergSchneider2021Sims4ADL} as a training dataset. For testing, we use real data, which originates from the Toyota Smarthome~\cite{das2019toyota} and ETRI~\cite{jang2020etriactivity3d} datasets.
Sims4Action provides ten hours of synthetic video material recorded from the computer game Sims 4, covering 10 basic human actions of daily living which have direct correspondences in the two real datasets.
Toyota Smarthome~\cite{das2019toyota} contains video material of 18 subjects performing 31 unscripted activities of daily living within a single apartment, and ETRI~\cite{jang2020etriactivity3d} is composed of 50 subjects performing 55 actions recorded from various perspectives that robots can be located in a home environment. However, we use only the 10 action correspondences to Sims4Action from both real datasets for our evaluation. We use the official cross-subject test split from Toyota Smarthome~\cite{dai2020toyota, das2019toyota}, in which 7 of the subjects are reserved for testing. The same protocol is adopted in the state-of-the-art results~\cite{RoitbergSchneider2021Sims4ADL} on the \textsc{Synthetic$\rightarrow$Real} benchmark. For ETRI, we use the unsupervised domain adaptation protocol from~\cite{kim2021action} and use the whole dataset for testing. Note that during test-time we extract the pose and optical flow modalities from the real test samples so that they match the input format of the $AC$ network.

As our evaluation data was collected in real households~\cite{das2019toyota, jang2020etriactivity3d}, the number of samples per activity class is unbalanced and follows a Zipf-like distribution rather than a uniform one.
We therefore focus on the \textit{balanced accuracy} (mean per-class accuracy) as our main evaluation metric and additionally report the \textit{unbalanced accuracy} (correct prediction rate among the complete test set). 
Note, that while we also evaluate the unbalanced accuracy to be consistent with ~\cite{RoitbergSchneider2021Sims4ADL}, the metric is highly biased towards overrepresented categories in the test set and should be taken with caution especially in real-life datasets, where the categories are not evenly distributed. 
We therefore consider the balanced accuracy as a much more reliable and less biased metric, as per-class-averaged metrics are used in most of the unbalanced activity recognition datasets~\cite{das2019toyota, caba2015activitynet,MartinRoitberg2019}.

\subsection{Qualitative Analysis}

First, we inspect the domain generation results produced by the domain generator $DG$ provided in Figure \ref{fig:modality_examples}. 
The model has learned to produce novel appearances, which might seem unusual at first sight, but the activity semantics are preserved through the additional classification loss $\mathcal{L}_{Class}$. Note, that the movement typical for the activity is more obvious in video results, than in still images.

To better understand the learned representations, we visualize the output of the domain generator's bottleneck layer with the t-distributed stochastic neighbor embeddings (tSNE) \cite{van2008visualizing} in Figure \ref{fig:tsne}.
The embeddings are produced by sampling 10 random frames from each video from the Sims4Action dataset for each of the source modalities. Each image sample is fed to the domain generator and its embedding is used for the t-SNE visualization. The $DG$ model is then applied again to the novel modalities $\widetilde{X}_k$ to obtain their embeddings. The plot in Figure \ref{fig:tsne} shows that each modality forms an individual cluster of points. This confirms that the generator has learned features which help differentiate between all 8 modalities $X_k \cup \widetilde{X}_k$. Furthermore, the modalities span a wider and non-overlapping distribution in the latent space, i.e. the training dataset has been diversified by adding the novel modalities $\widetilde{X}_k$, which leads to better domain generalization. 
%
%
%
%
\subsection{Quantitative Results} \label{subsec:quantitative}

In Table \ref{tab:results_accuracy}, we compare the $AC$s trained only on the source modalities $X_k$ to the ones which are trained including the generated novel modalities $\widetilde{X}_k$. All modalities $X_k \cup \widetilde{X}_k$ are produced from Sims4Action~\cite{RoitbergSchneider2021Sims4ADL} only. Additionally, we consider all 15 modality combinations and evaluate all models on the real Toyota Smarthome~\cite{das2019toyota} and ETRI~\cite{jang2020etriactivity3d} datasets via the balanced and unbalanced accuracy metrics. 

For most combinations, the extension of the training dataset with novel modalities increases the balanced accuracy significantly, in some cases by up to 13\% points (ETRI: L). 
This is especially prevalent in multi-modal settings joined over early fusion, and we hypothesize that the generation of additional data alleviates overfitting problems which arise when training a larger, multi-modal model. Optical flow (OF) profits the least from this technique when combined with other modalities, in some cases performing worse than the baseline. This might be explained by OF's low accuracy on the \textsc{Synthetic$\rightarrow$Synthetic} benchmark (44\%), i.e. OF is a weak modality on Sims4Action. However, performance losses in these cases are small in comparison to the significant gains which are achieved by the other combinations. We do list unbalanced accuracy as well, which provides similar results. Note that the improvement depends on the modality combination and test dataset. However, the average improvement for the balanced accuracy for Toyota Smarthome/ETRI is +3.1\%/+2.8\% and +4.7\%/+8.0\% for unbalanced accuracy.



\begin{table}[!h]
	\centering
	\scalebox{0.8}{
		\begin{tabular}{@{}ll|cc|cc@{}}
			\toprule
			\makecell{Approach} & \makecell{Pre-training} &
			\multicolumn{2}{c}{Toyota SmartHome} & \multicolumn{2}{c}{Etri Activity}\\ \hline
			{} & {} & 			\makecell{Balanced \\ Accuracy}& \makecell{Unbalanced \\ Accuracy} &
		    \makecell{Balanced \\ Accuracy}& \makecell{Unbalanced \\ Accuracy}\\
			\midrule
			Random Choice & None & 10.00 & 10.00 & 10.00 & 10.00\\
	
	        \hline
            \multicolumn{6}{l}{\cellcolor{gray!10}\noindent\textbf{Currently reported state-of-the-art results on the \textsc{Synthetic$\rightarrow$Real} benchmark}} \\ 
	        \hline
	        S3D \cite{carreira2017quo, RoitbergSchneider2021Sims4ADL} & None & 12.40 & 19.95 & 11.71 & 13.86 \\ 
	        S3D \cite{carreira2017quo, RoitbergSchneider2021Sims4ADL} & Kinetics-400 \cite{kay2017kinetics} & 23.25 & 22.75 & 23.45 & 28.57 \\ \hline
	        \multicolumn{6}{l}{\cellcolor{gray!10}\noindent\textbf{Other domain generalization methods on the \textsc{Synthetic$\rightarrow$Real} benchmark}} \\ 
	        \hline
		    TA$^3$N \cite{chen2019temporal} & ImageNet \cite{deng2009imagenet} & 14.19 & 14.44 & 25.11 & 35.12 \\ 
		    APN \cite{yao2021videodg} & ImageNet \cite{deng2009imagenet} & 22.09 & 17.19 & 27.97 & 37.67 \\ 
		    VideoDG \cite{yao2021videodg} & ImageNet \cite{deng2009imagenet} & 25.71 & 21.12 & 25.55 & 41.09 \\ \hline
		    
            \multicolumn{6}{l}{\cellcolor{gray!10}\textbf{Our model with domain generation (best modality combination)}} \\ 
	        \hline
		    Ours  & None & \textbf{27.73}  & \textbf{32.20} & \textbf{29.05} & \textbf{41.88} \\

			\bottomrule
		\end{tabular}
	}
	
	\caption{Comparison of our model to state-of-the-art results on the \textsc{Synthetic$\rightarrow$Real} benchmark~\cite{RoitbergSchneider2021Sims4ADL} and to three additional domain generalization methods - TA$^3$N~\cite{chen2019temporal}, APN~\cite{yao2021videodg}, and VideoDG~\cite{yao2021videodg}.}
	
	\label{table:soa}
	\vspace{-1em}
\end{table}

\begin{table*}[t]
\centering
\scalebox{0.80}{
\begin{threeparttable}
 \begin{tabular}[t]{ l | c l | c l | c l | c l | c c}
\toprule
{} & \multicolumn{2}{c}{\thead{ \textbf{Balanced Accuracy}}} & \multicolumn{2}{c|}{\thead{ \textbf{Unbalanced Accuracy}}} & \multicolumn{2}{c}{\thead{ \textbf{Balanced  Accuracy}}} & \multicolumn{2}{c|}{\thead{ \textbf{Unbalanced Accuracy}}} & \multicolumn{1}{c}{\thead{ \textbf{Bal. Acc.}}} & \multicolumn{1}{c}{\thead{ \textbf{Unbal. Acc.}}} \\ \hline
%
{} & \multicolumn{8}{c|}{\thead{\textsc{Synthetic$\rightarrow$Real}}} & \multicolumn{2}{c}{\thead{\textsc{Synthetic$\rightarrow$Synthetic}}} \\ \hline


\multicolumn{1}{l|}{\cellcolor{gray!10}\textbf{Test Set}} & \multicolumn{4}{c|}{\cellcolor{gray!10}{\textbf{Toyota Smarthome}~\cite{das2019toyota}}} & \multicolumn{4}{c|}{\cellcolor{gray!10}{\textbf{ETRI}~\cite{jang2020etriactivity3d}}} & \multicolumn{2}{c}{\cellcolor{gray!10}{ \textbf{Sims4Action}~\cite{RoitbergSchneider2021Sims4ADL}}} \\ \hline
\multirow{1}{*}{\thead{\diagbox[height=2.7em]{Input}{Train Set}}} 
&\multirow{2}{*}{\thead{Only $X_k$}}&\multirow{2}{*}{\thead{Ours: $X_k \cup \widetilde{X}_k$}} & \multirow{2}{*}{\thead{Only $X_k$}}&\multirow{2}{*}{\thead{Ours: $X_k \cup \widetilde{X}_k$}} & \multirow{2}{*}{\thead{Only $X_k$}}&\multirow{2}{*}{\thead{Ours: $X_k \cup \widetilde{X}_k$}} & \multirow{2}{*}{\thead{Only $X_k$}}&\multirow{2}{*}{\thead{Ours: $X_k \cup \widetilde{X}_k$}} &
\multicolumn{2}{c}{\multirow{2}{*}{\thead{Only $X_k$}}}\\
[2em]
\hline
\multicolumn{11}{l}{\cellcolor{gray!10}}\\[-0.8em]
\multicolumn{11}{l}{\cellcolor{gray!10} \textbf{Individual}} \\
\hline
RGB &  $13.7$ & $13.3$ \deltam{0.4} & $18.5$ & $17.6$ \deltam{0.9} & $11.7$ & $15.0$ \deltap{3.3} & $13.9$ & $15.6$ \deltap{1.7} & $61.8$ & $59.4$ \\
Heatmaps (H) & $20.2$ & $25.8$ \deltap{5.6} & $20.0$ & $23.7$ \deltap{3.7} & $15.2$ & $17.9$ \deltap{2.7} & $22.2$ & $40.9$ \deltap{18.7} & $71.4$ & $70.4$ \\
Limbs (L) & $22.0$ & \textbf{27.7} \deltap{5.7} & $21.9$ & $21.7$ \deltam{0.2} & $16.1$ & \textbf{29.1} \deltap{13.0} & $16.2$ & $38.7$ \deltap{22.5} & $75.1$ & $74.4$ \\
Optical Flow (OF) & $21.3$ & $22.6$ \deltap{1.3} & $31.5$ & \textbf{32.2} \deltap{0.7} & $11.6$ & $14.7$ \deltap{3.1} & $13.5$ & $17.5$ \deltap{4.0} & $44.5$ & $43.7$\\
\hline
\multicolumn{11}{l}{\cellcolor{gray!10}}\\[-0.8em]
\multicolumn{11}{l}{\cellcolor{gray!10}\textbf{Early Fusion by Channel Concatenation}} \\
\hline
RGB + H & $10.3$ & $20.7$ \deltap{10.4} & $6.1$ & $22.8$ \deltap{16.7} & $7.3$ & $12.9$ \deltap{5.6} & $7.6$ & $14.8$ \deltap{7.2} & $76.8$ & $77.2$ \\
RGB + L & $13.5$ & $18.5$ \deltap{5.0} & $8.2$ & $18.6$ \deltap{10.4} & $13.4$ & $15.7$ \deltap{2.3} & $15.9$ & $23.5$ \deltap{7.6} & $81.5$ & $81.0$ \\
RGB + OF & $19.4$ & $17.0$ \deltam{2.4} & $31.0$  & $30.4$ \deltam{0.6} & $14.6$ & $14.1$ \deltam{0.5} & $20.9$ & \textbf{41.9} \deltap{21.0} & $70.4$ & $72.3$ \\
\hline
H + L & $15.7$ & $25.3$ \deltap{9.6}& $15.7$  & $20.2$ \deltap{4.5} & $15.0$ & $17.8$ \deltap{2.8} & $16.5$ & $17.9$ \deltap{1.4} & $57.8$ & $50.1$ \\
H + OF & $16.7$ & $23.7$ \deltap{7.0} & $20.0$  & $24.6$ \deltap{4.6} & $12.6$ & $14.2$ \deltap{1.6} & $15.7$ & $17.9$ \deltap{2.2} & $71.1$ & $64.7$ \\
L + OF & $19.7$ & $19.8$ \deltap{0.1} & $24.8$  & $22.1$ \deltam{-2.7} & $11.4$ & $12.5$ \deltap{1.1} & $13.4$ & $16.7$ \deltap{3.3} & $72.7$ & $71.2$ \\
\hline
RGB + H + L & $15.3$ & $13.9$ \deltam{1.4}  & $10.0$ & $23.2$ \deltap{13.2} & $5.8$ & $12.2$ \deltap{6.4} & $5.8$ & $13.8$ \deltap{8.0} & $73.7$ & $73.1$ \\
RGB + H + OF & $11.6$ & $13.4$ \deltap{1.8}  & $13.0$ & $29.5$ \deltap{16.5} & $12.8$ & $14.2$ \deltap{1.4} & $12.5$ & $22.3$ \deltap{9.8} & $70.7$ & $73.1$ \\
RGB + L + OF & $12.0$ & $14.1$ \deltap{2.1}  & $15.0$ & $13.5$ \deltam{1.5} & $11.5$ & $10.1$ \deltam{1.4} & $15.9$ & $17.8$ \deltap{1.9} & $83.0$ & $79.8$ \\
\hline
H + L + OF & $25.1$ & $20.7$ \deltam{4.4} & $36.8$  & $22.9$ \deltam{13.9} & $15.2$ & $12.5$ \deltam{2.7} & $13.9$ & $15.6$ \deltap{1.7} & $77.1$ & $72.9$ \\
\hline
RGB + H + L + OF & $12.1$ & $18.6$ \deltap{6.5} & $7.0$ & $27.4$ \deltap{20.4} & $10.3$ & $13.0$ \deltap{2.7} & $9.1$ & $18.9$ \deltap{9.8} & $66.8$ & $65.1$ \\
\bottomrule
\end{tabular}
\end{threeparttable}
}
\caption{Evaluation results for all $15$ modality combinations on the \textsc{Synthetic$\rightarrow$Real} benchmark for the real Toyota Smarthome~\cite{das2019toyota} and ETRI~\cite{jang2020etriactivity3d} datasets and on the \textsc{Synthetic$\rightarrow$Synthetic} benchmark for Sims4Action~\cite{RoitbergSchneider2021Sims4ADL}. Our method of generating novel modalities $\widetilde{X}_k$ for the training data leads to a significant improvement of the vast majority of the models on the \textsc{Synthetic$\rightarrow$Real} benchmark. The performance on Sims4Action is listed to illustrate the domain gap. The performance boost by adding novel modalities $\widetilde{X}_k$ to the training data is indicated in brackets for each modality combination.}
\label{tab:results_accuracy}
\vspace{-1em}
\end{table*}



\subsection{Comparison to State-of-the-art and Other Approaches}
In Table \ref{table:soa} we compare the best results achieved by our approach (\textit{i.e.}, the best modality variants from Table \ref{tab:results_accuracy} marked in bold), and compare them with current state-of-the-art results on the Sims4Action \textsc{Synthetic$\rightarrow$Real} benchmark \cite{RoitbergSchneider2021Sims4ADL}. We also compare our method to three other domain generalization methods: TA$^3$N~\cite{chen2019temporal}, APN~\cite{yao2021videodg}, and VideoDG~\cite{yao2021videodg}. To this end, we train all three models end-to-end on Sims4Action~\cite{RoitbergSchneider2021Sims4ADL} and follow the same evaluation protocols for Toyota Smarthome~\cite{das2019toyota} and ETRI~\cite{jang2020etriactivity3d}, which we used for our approach in Section \ref{subsec:quantitative}.

Despite not making use of pre-training, we significantly improve the state-of-the-art on both accuracy metrics and on both real test datasets. Our method also consistently outperforms the other domain generalization methods TA$^3$N~\cite{chen2019temporal}, APN~\cite{yao2021videodg}, and VideoDG~\cite{yao2021videodg}. APN and VideoDG also outperform the previous state-of-the-art on some metrics, which supports the claim that a domain generalization strategy for the \textsc{Synthetic$\rightarrow$Real} benchmark is beneficial.

\section{Conclusion and Limitations}

In this paper, we explored the paradigm of novel domain generation for recognizing Activities of Daily Living (ADL).
Our work is motivated by the idea, that such synthesis of novel action appearances diversifies the training data and therefore mitigates the problem of domain shift. 
We specifically aim for \textsc{Synthetic$\rightarrow$Real} ADL recognition and introduce a multimodal framework which leverages RGB, body pose, joint heatmaps and optical flow to learn generating novel modalities. Our experiments confirm that complementing training data with novel modalities leads to significant improvements in domain generalization, outperforming previous state-of-the-art results on the \textsc{Synthetic$\rightarrow$Real} benchmark and other domain generalization approaches on both real test datasets Toyota Smarthome and ETRI. 

While our method strongly improves the performance in case the data appearance has changed at test-time, it is not without limitations. 
First, we observe, that the model complexity increases with the number of source modalities, leading to a difficult optimization and longer training times. Secondly, while we achieve state-of-the-art results on the  \textsc{Synthetic$\rightarrow$Real} benchmark, we acknowledge, that this comes with additional computational cost, as the body pose needs to be estimated first.
Nevertheless, our work makes a step towards real-life utilization of synthetic datasets, which would enable far less intrusive data collection for raising action recognition capabilities of ADL robotic applications.

\mypar{Acknowledgements} This work was supported by the JuBot project sponsored by the Carl Zeiss Stiftung and Competence Center Karlsruhe for AI Systems Engineering (CC-KING) sponsored by the
Ministry of Economic Affairs, Labour and Housing Baden-W{\"u}rttemberg.


{\small
	\bibliographystyle{bibtex/IEEEtran}
	\balance
	\bibliography{egbib}
}

\end{document}